# Efficient Graph Encoder Embedding for Large Sparse Graphs in Python

Xihan Qin
University of Delaware, Newark, USA, xihan@udel.edu, ORCID: 0000-0003-1096-5941

Cencheng Shen
University of Delaware, Newark, USA, shenc@udel.edu, ORCID: 0000-0003-1030-1432

**Abstract:** Graph is a ubiquitous representation of data in various research fields, and graph embedding is a prevalent machine learning technique for capturing key features and generating fixed-sized attributes. However, most state-of-the-art graph embedding methods are computationally and spatially expensive. Recently, the Graph Encoder Embedding (GEE) is shown as the fastest graph embedding technique and is suitable for a variety of network data applications. As real-world data often involves large and sparse graphs, the huge sparsity usually results in redundant computations and storage. To address this issue, we propose an improved version of GEE, sparse GEE, which optimizes the calculation and storage of zero entries in sparse matrices to enhance the running time further. Our experiments demonstrate that the sparse version achieves significant speedup compared to the original GEE with Python implementation for large sparse graphs, and sparse GEE is capable of processing millions of edges within minutes on a standard laptop.

## 1. INTRODUCTION

Graph is a unique and versatile data structure that captures relationships among objects. A graph G, denoted by G = {V, E}, consists of a set of nodes V and edges E, where nodes represent objects and edges represent connections between them. Additional information, such as node attributes and edge weights, can also be incorporated into the graph. Graphs are used in a variety of research areas, such as protein-protein interaction networks [1], social networks [2], document citations [3], and recommender systems [4], among others.

Graph embedding methods [5] are a popular class of methods that capture the significant parts of the graph connections into a low-dimensional Euclidean space. There are two primary types of graph embedding methods: one that transforms the entire graph into a single vector and another that transforms the graph into a set of vectors, with each vector representing a node [6]. Some popular examples of the latter type include spectral embedding [7], node2vec[8], graph convolutional neural network [9], and most recently, graph encoder embedding (GEE) [10]. GEE outputs a final embedding matrix Z with a size of N × K, where K represents the number of known classes.

Most real graphs are sparse by nature, which leads to significant overhead in running time and storage. GEE [10] was designed with efficiency and speed in mind, utilizing an edge list that only stores connected nodes and thereby avoids computation and storage on zero entries in the adjacency matrix. Furthermore, GEE has demonstrated remarkable versatility in its application to various graph-related tasks, including graph bootstrap, vertex classification, vertex clustering, community detection, time-series graph and multiple-graph inference [10-13].

In this study, we aim to enhance the efficiency and storage capacity of the GEE method by implementing special data structures for sparse matrices in both intermediate and final stages, along with three additional embedding options: Laplacian normalization, diagonal augmentation, and correlation. Our proposed method, sparse GEE, can be easily executed on a standard laptop. Our experiments demonstrate that with all options activated, sparse GEE outperforms GEE, achieving a 2.5 times speedup on our largest real dataset (consisting of 600 thousand nodes and 10 million edges) and an 86 times speedup on our largest simulated dataset (with 10 thousand nodes and 5.6 million edges). These results highlight the practical usefulness of sparse GEE in handling large, sparse graphs efficiently. The source code is available in our GitHub repository at https://github.com/xihan-qin/GEE_sparse.

## 2. PREVIOUS WORK

GEE [10] embeds the graph using the label information and local connections. The vertex labels can be either true classification information or predicted labels. The label counts are used to generate an intermediate weight matrix W of size N × K. Each row of W is a weight vector of the corresponding node, and for any node j belonging to class k, its weight vector $W_j$ is [0 … $1/n_k$ … 0], with $n_k$ denotes the total node count of class k. The matrix W can be considered as a normalized one-hot encoding, where 1 is replaced by $1/n_k$.

The embedding matrix Z is obtained by multiplying the adjacency matrix A with W:

$$Z = AW \quad (1)$$

Here, Z has a size of N × K, where each row represents a vector of the corresponding node in a latent space, and K is the number of classes. The embedding vector $Z_i$ for node i is [$Z_i^0$ … $Z_i^k$ … $Z_i^{K-1}$] with length K. For each node i and class k, the embedding value $Z_i^k$ is calculated as follows:

$$Z_i^k = \sum_{j \in S_{ik}} e_{ij} W_j^k$$

Here, $S_{ik}$ denotes the neighboring nodes of i that are labeled as class k, $e_{ij}$ is the edge weight, and $W_j^k$ denotes the weight value picked from vector $W_j$ for class k.

GEE has a linear time complexity and is faster than other commonly used embedding methods [10]. The algorithm employs the edge list to achieve linear complexity, which is computationally efficient and requires less storage compared to the adjacency matrix. Each row of the edge list contains three elements: node i, node j, and eij, where eij represents the edge weight between node i and node j. In the absence of edge weight information, all edges are assigned a weight of 1. The algorithm was implemented in MATLAB, Python, and R.

GEE offers 3 additional options: diagonal augmentation, Laplacian normalization, and correlation. The diagonal augmentation option adds self-connection to all the nodes of the graph before calculating the embedding matrix, replacing A in Eq. (1) with (A + I), where I denotes the identity matrix. The Laplacian normalization option performs the edge list version of Laplacian normalization before calculating the embedding matrix, replacing A in Eq. (1) with $D^{-0.5}AD^{-0.5}$, where D denotes the degree matrix. The correlation option performs 2-norm on each row of the embedding matrix Z, and the normalized matrix Z' becomes the final embedding matrix.

## 3. METHOD

In practice, most graphs are both large and sparse, which can lead to a significant number of zero entries if represented by an adjacency matrix [14]. A matrix with a large number of zero entries is referred to as a sparse matrix [15]. Operating directly on the sparse matrices for large graphs requires a large amount of space and time to store. However, zero entries do not affect most matrix calculations and have little significance to store. To address this issue, the original GEE algorithm uses the edge list instead of the adjacency matrix to compute the embedding matrix, which significantly improves running time [10]. However, the resulting embedding matrix remains sparse, and a better data structure is needed to store it more efficiently. Intermediate steps in the GEE algorithm also involve sparse matrices. For example, the degree matrix and identity matrix used in Laplacian normalization and diagonal augmentation options are diagonal matrices with non-zero values only on the diagonal and zeros elsewhere. Additionally, the intermediate weight matrix W in the original GEE method is also very sparse, with only one weight value of $1/n_k$ for each node belonging to class k and zeros for all other classes. Using a better data structure to store these sparse matrices can save a lot of space and unnecessary operations involving zero entries.

In this study, we present sparse GEE, which improves the original GEE in two aspects. Firstly, sparse GEE uses the Compressed Sparse Row (CSR) data structure [16] to calculate the embedding matrix, including for the additional options. Secondly, when constructing intermediate results, sparse GEE uses the Dictionary of Keys (DOK) data structure, which is then transformed to CSR format for further calculations and operations. Fig.1 provides an example of a sparse matrix and its corresponding edge list and CSR formats. Both edge list and CSR save space compared to the original weighted adjacency matrix (the example sparse matrix), as they exclude zero entries entirely. In CSR, there are 3 lists: col_indices, data, and index_pointers. For the purpose of this analysis, we will use E to denote the number of edges, and R for the number of rows. Starting from the first row and proceeding to the last row, "col_indices" stores the column indices that have a data value in the matrix, while "data" stores the corresponding values in the matrix in the same order. The indices of "index_pointers", except for the last index, correspond to the row indices in the original sparse matrix. The values in "index_pointers" represent the pointers for the start index (index_pointers [row_i]) and the end index (index_pointers [row_i+1]) for both "col_indices" and "data" for each row row_i. For instance, let us take row_2 in the adjacency matrix, which has a value of 2 at col_1 and a value of 3 at col_5. The row index in the matrix is 2, so the start and end pointers for row_2 are 3 and 5, which are index_pointers [2] and index_pointers [3] respectively. Then, we can retrieve the matrix column indices 1 and 5 by accessing col_indices [3:5], and the corresponding data values 2 and 3 by accessing data [3:5]. The length of "index_pointers" is (R+1) because it needs to include an additional pointer for the end of the last row, which is row_r + 1, where row_r is the index of the last row. Therefore, compared to the 3 × E size of the edge list, CSR requires less space as long as E > R + 1. The larger and sparser the matrix, the more significant the advantage of using CSR, which is typically the case for most real-world graph data analysis.

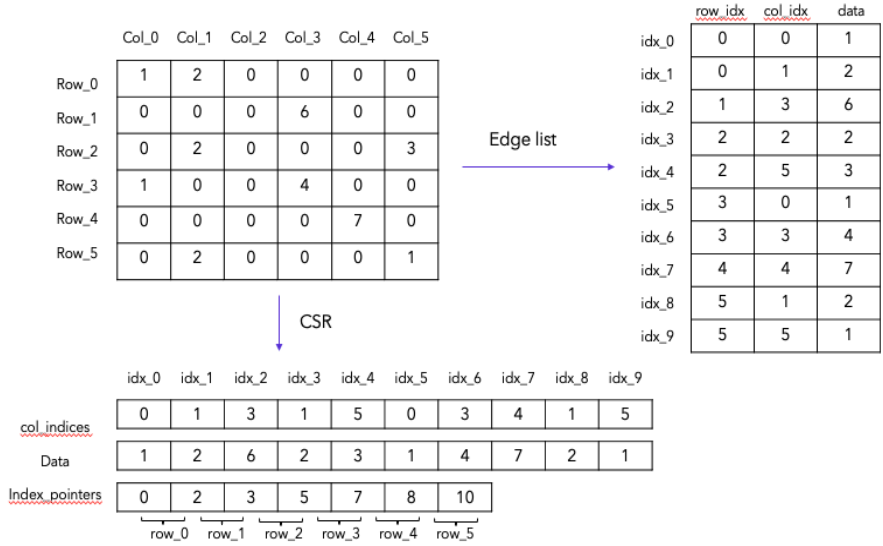

*Figure 1. An example that compares edge list and CSR in sparse matrix storage.*

Table 1 lists the formulas for embedding matrix $Z_S$ or $Z'_S$ when different options are selected. The subscript s in $Z_S$ or $Z'_S$ indicates the use of CSR, and the same notation applies to other variables: $A_S$ and $W_S$ represent the adjacency matrix and intermediate weight matrix in CSR format, respectively, while $I_S$ and $D_S$ represent the identity matrix and degree matrix in diagonal CSR format respectively.

*Table 1. Formulas in Sparse GEE*

| Sparse GEE with different options | Formula of Embedding matrix |
| --- | --- |
| Sparse GEE | $Z_S = A_S W_S$ |
| Sparse GEE + diagonal augmentation | $Z_S = (A_S + I_S)W_S$ |
| Sparse GEE + Laplacian normalization | $Z_S = (D_S^{-0.5} A_S D_S^{-0.5})W_S$ |
| Sparse GEE + correlation | $Z'_S = \dfrac{A_S W_S}{(\|(A_S W_S)^T\|_2)^T}$ |

The implementation of data structures for sparse matrices and their calculations in this project use the SciPy 2-D sparse array package (scipy.sparse) [17]. The Python implementation of sparse GEE can be accessed at https://github.com/xihan-qin/GEE_sparse. Additionally, we provide code for users to convert the edge list to a sparse matrix.

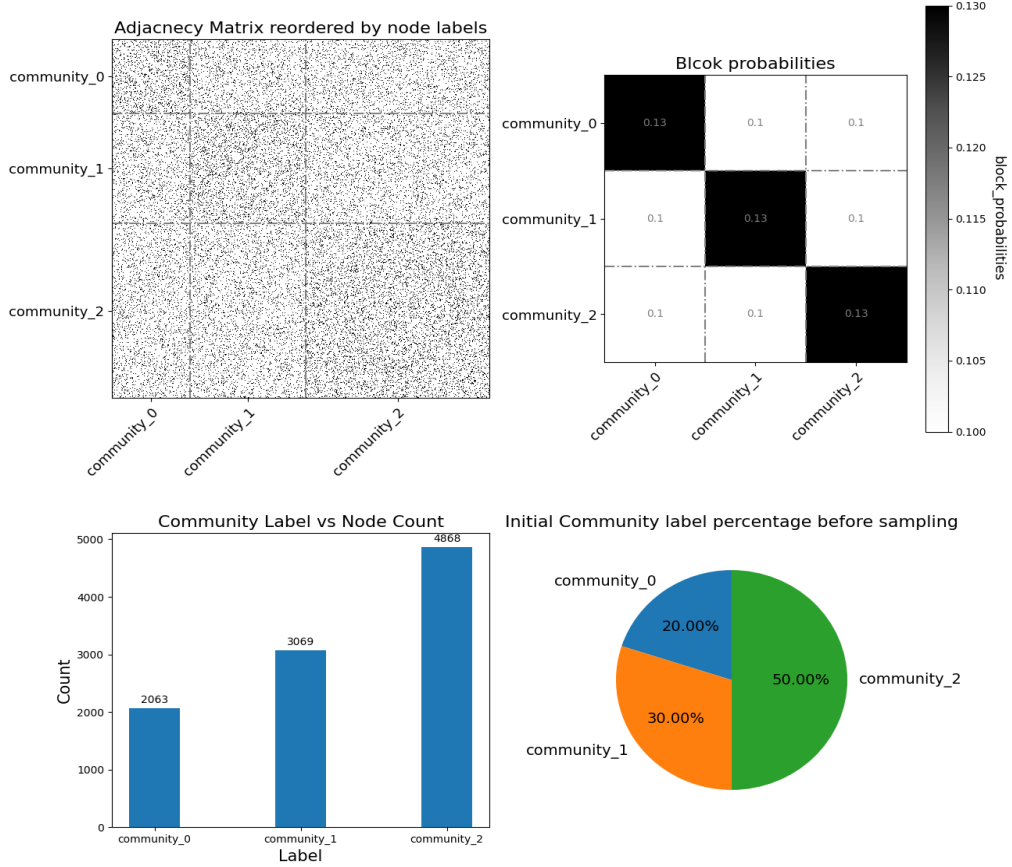

*Figure 2. SBM with node size 10,000.*

## 4. EVALUATIONS

In this study, we compared the original GEE with sparse GEE by simulating graphs using the Stochastic Block Model (SBM) [18-19]. The SBM graphs were generated with 3 classes having class probabilities [0.2, 0.3, 0.5], between-class probability 0.1, and within-class probability 0.13. The only difference between the graphs was in the number of nodes, which were set to 100, 1000, 3000, 5000, and 10,000. Fig.2 shows the SBM graph with 10,000 nodes, where the upper left plot displays the density of the classes (also known as communities or blocks); the upper right plot shows the block probabilities used for generating edges; the lower left plot shows the label count and node count for each class, and the lower right plot shows the percentage of each class in the population.

*Table 2. Benchmark Datasets*

| Datasets | Nodes | Edges | Classes | Edge Density ($d$) |
|---|---|---|---|---|
| Citeseer | 3,327 | 4,732 | 6 | 0.00085 |
| Cora | 2,708 | 5,429 | 7 | 0.00148 |
| Proteins-all | 43,471 | 162,088 | 3 | 0.00017 |
| PubMed | 19,717 | 44,338 | 3 | 0.00023 |
| CL-100K-1d8-L9 | 92,482 | 373,986 | 9 | 0.00009 |
| CL-100K-1d8-L5 | 92,482 | 10,000,000 | 5 | 0.00234 |

We also analyzed real-world data using benchmark datasets from Network Depository [20]. Table 2 provides details about each dataset, including Edge Density, which serves as a measure of the graph's sparsity. Edge Density d is calculated using Eq. (2), where |E| represents the number of edges and |V| represents the number of nodes.

$$d = \frac{2|E|}{|V| \times (|V| - 1)} \qquad (2)$$

All tests described below were performed on a laptop with a 2 GHz Quad-Core Intel Core i5 processor and 16 GB 3733 MHz LPDDR4X memory, and the algorithm was implemented using Python language.

## 4.1. Simulated Datasets:

We evaluated the performance of GEE and sparse GEE on a range of simulated datasets, varying in size from 0.1 to 10 thousand nodes, with edges counts ranging from 0.6 thousand to 5.6 million. Fig.3 displays the running time for both algorithms with all additional options enabled (Lap = T, Diag = T, Cor = T). It is evident that sparse GEE scales up much better than GEE as the graph size increases. For the largest simulated graph with 10 thousand nodes and 5.6 million edges, sparse GEE only takes an average of 0.6 seconds, while GEE takes 52.4 seconds, resulting in sparse GEE being 86 times faster than GEE.

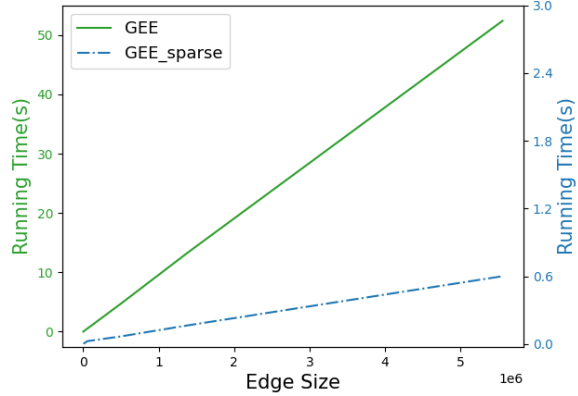

*Figure 3. Compare GEE with sparse GEE on simulated datasets (SBM)*

## 4.2. Real Datasets:

We conducted tests and compared GEE and sparse GEE on the real datasets with all possible settings on the additional options (Laplacian, diagonal, and correlation). The comparison results for these datasets are shown in Tables 3-4. Sparse GEE can embed the largest sparse real dataset "CL-100K-1d8-L5" with 0.6 million nodes and 10 million edges using all three additional options in 174.552 seconds on a single laptop, whereas GEE takes 604.018 seconds.

Laplacian normalization is the most computationally expensive among the three options, as it requires more operations to calculate the embedding than the other options. Specifically, the algorithm needs to construct the degree matrix, power the degree matrix, and finally calculate the dot product (as shown in Table 1). On the other hand, for the diagonal option (Table 1), we only need to add one to the diagonal before the calculation. For the correlation option, we just need to first calculate the row's norm and then normalize each entry by the norm.

Our results indicate that sparse GEE with the Laplacian option achieved significantly better performance (up to 4 times faster) than the original GEE on all test cases. When the Laplacian option is turned off, the original GEE may be faster than sparse GEE on some small graphs. However, as the graphs get larger, sparse GEE consistently shows much better performance than the original GEE.

*Table 3. GEE vs Sparse GEE on Real Datasets with different settings on the additional options (1)*

| Data Set (node/edge) | Operation time (s) | | | | | | | |
|---|---|---|---|---|---|---|---|---|
| | Lap = T, Diag = T, Cor = T | | Lap = T, Diag = T, Cor = F | | Lap = T, Diag = F, Cor = T | | Lap = T, Diag = F, Cor = F | |
| | GEE | Sparse GEE | GEE | Sparse GEE | GEE | Sparse GEE | GEE | Sparse GEE |
| CiteSeer (3264/4536) | 0.097 | **0.032** | 0.097 | **0.032** | 0.053 | **0.032** | 0.051 | **0.030** |
| Cora (2708/5429) | 0.088 | **0.027** | 0.088 | **0.027** | 0.068 | **0.033** | 0.068 | **0.026** |
| proteins-all (43471/162088) | 2.259 | **0.419** | 2.259 | **0.419** | 1.866 | **0.391** | 1.846 | **0.478** |
| PubMed (19717/44338) | 0.739 | **0.201** | 0.739 | **0.201** | 0.673 | **0.208** | 0.560 | **0.199** |
| CL-100K-1d8-L9 (92482/373986) | 4.850 | **0.978** | 4.850 | **0.978** | 4.166 | **0.992** | 3.823 | **1.095** |
| CL-100K-1d8-L5 (92482/10000000) | 604.018 | **174.552** | 585.288 | **147.790** | 633.746 | **118.705** | 571.360 | **123.691** |

*Table 4. GEE vs Sparse GEE on Real Datasets with different settings on the additional options (2)*

| Data Set (node/edge) | Operation time (s) | | | | | | | |
|---|---|---|---|---|---|---|---|---|
| | Lap = F, Diag = T, Cor = T | | Lap = F, Diag = T, Cor = F | | Lap = F, Diag = F, Cor = T | | Lap = F, Diag = F, Cor = F | |
| | GEE | Sparse GEE | GEE | Sparse GEE | GEE | Sparse GEE | GEE | Sparse GEE |
| CiteSeer (3264/4536) | **0.024** | 0.031 | **0.024** | 0.033 | **0.016** | 0.034 | **0.014** | 0.031 |
| Cora (2708/5429) | **0.024** | 0.025 | **0.026** | 0.026 | **0.023** | 0.024 | **0.019** | 0.025 |
| proteins-all (43471/162088) | 0.830 | **0.399** | 0.623 | **0.411** | 1.143 | **0.432** | 0.518 | **0.462** |
| PubMed (19717/44338) | 0.231 | **0.201** | 0.228 | **0.185** | **0.170** | 0.188 | **0.177** | 0.183 |
| CL-100K-1d8-L9 (92482/373986) | 1.330 | **0.909** | 1.322 | **0.936** | 1.114 | **0.924** | **1.058** | 1.360 |
| CL-100K-1d8-L5 (92482/10000000) | 203.860 | **108.977** | 192.780 | **132.160** | 171.838 | **125.935** | 171.714 | **106.264** |

## 5. CONCLUSION

In this work, we introduced sparse GEE, an improved version of the GEE algorithm that addresses both space and speed limitations when dealing with large sparse graphs. Our experiments on simulated datasets demonstrate that sparse GEE scales significantly better than GEE as the graph size increases, achieving up to 84 times faster performance on the SBM graph with 5.6 million edges. Results on real datasets also show that sparse GEE outperforms GEE on large sparse graphs, especially when the Laplacian option is enabled.

It is worth mentioning that the original GEE algorithm is already faster than other state-of-the-art graph embedding algorithms. When the dataset is small and no additional options are required, using the original GEE may be slightly faster. This is because GEE calculates the embedding using the edge list of the graph and the constructed matrix W directly, and the edge list already avoids null connections. In contrast, sparse GEE requires constructing a sparse weight matrix $W_S$ using DOK format, transforming DOK into CSR format, and then performing the calculation. Therefore, when the graphs are small, the time required for constructing and transforming the sparse format in sparse GEE can outweigh the benefits of using a sparse format to calculate the embedding matrix and store intermediate matrices.

When dealing with large and sparse graphs, which is often the case in practice, sparse GEE further improves the GEE algorithm in terms of running time and storage. Sparse GEE avoids wasting storage and computation on zero entries for all matrices involved, whereas the original GEE only avoids storage and computation on the adjacency matrix (the benefit of using the edge list directly). Therefore, sparse GEE is a better option for embedding large graphs. For small graphs or when no additional options are required, the original GEE algorithm can sometimes be considered as an alternative.

## ACKNOWLEDGMENT

This work was supported in part by the National Science Foundation DMS-2113099, and by funding from Microsoft Research.

**Xihan Qin** is a Ph.D. student in Computer Science at the University of Delaware. She received the MS degree in Biotechnology from Georgetown University in 2015 and the MS degree in Bioinformatics from the University of Delaware in 2021. Her research interests include graph machine learning, bioinformatics, and computational biology.

**Cencheng Shen** is an Associate Professor in the Department of Applied Economics and Statistics at the University of Delaware. In addition, he holds joint appointment in the Data Science Institute and the Department of Mathematical Sciences. His research lies at the intersection of machine learning methods, statistical theory, and modern data analysis, with a particular focus on graph learning, neural networks, data fusion, dimension reduction, and hypothesis testing. His research projects have been funded by several prestigious organizations including NSF, DARPA, and Microsoft Research.